\PassOptionsToPackage{table}{xcolor}
\documentclass[letter, 10pt, runningheads]{llncs}
\usepackage{amsmath,amssymb,amsfonts}
\usepackage{url}

\usepackage[breaklinks,hidelinks]{hyperref}
\usepackage{algorithmic}
\usepackage{wrapfig}
\usepackage{caption}
\usepackage{subcaption}
\usepackage{tikz}
\usepackage{pgfplots}
\usepackage{graphicx}
\usepackage[ruled,vlined,linesnumbered]{algorithm2e}
\usepackage{multirow}
\usepackage{booktabs}
\usepackage[table]{xcolor}
\usepackage{verbatim}
\usepackage{enumitem}
\usepackage[margin=1.6in]{geometry} 
\usepackage[misc]{ifsym}
\usepackage{orcidlink}

\SetCommentSty{mycommfont}
\SetKwInOut{Parameter}{Input}

\usetikzlibrary{positioning, shapes.geometric, arrows.meta}
\def\BibTeX{{\rm B\kern-.05em{\sc i\kern-.025em b}\kern-.08em
    T\kern-.1667em\lower.7ex\hbox{E}\kern-.125emX}}

\newcommand{\fref}[1]{Fig.~\ref{#1}}

\newcommand{\sref}[1]{Section~\ref{#1}}

\begin{document}

\title{\Large \bf TSI-GAN: Unsupervised Time Series Anomaly Detection using Convolutional Cycle-Consistent Generative Adversarial Networks}
\titlerunning{TSI-GAN}
%

\author{Shyam Sundar Saravanan\inst{1}\and
Tie Luo \inst{1}\orcidlink{0000-0003-2947-3111}\textsuperscript{(\Letter)} \and Mao Van Ngo\inst{2}}
\authorrunning{S.S. Saravanan et al.}
%
\institute{Missouri University of Science and Technology, Rolla, MO 65401, USA \and
Singapore University of Technology and Design, Singapore 487372 \\
\{ssdmw, tluo\}@mst.edu, vanmao\_ngo@sutd.edu.sg}

\maketitle

\vspace{-25pt}
\begin{abstract}
Anomaly detection is widely used in network intrusion detection, autonomous driving, medical diagnosis, credit card frauds, etc. However, several key challenges remain open, such as 
lack of ground truth labels, presence of complex temporal patterns, and generalizing over different datasets. 
This paper proposes TSI-GAN, an unsupervised anomaly detection model for time-series that can learn complex temporal patterns automatically and generalize well, i.e., no need for choosing dataset-specific parameters, making statistical assumptions about underlying data, or changing model architectures. To achieve these goals, we convert each input time-series into a sequence of 2D images using two encoding techniques with the intent of capturing temporal patterns and various types of deviance. Moreover, we design a reconstructive GAN that uses convolutional layers in an encoder-decoder network and employs {\em cycle-consistency loss} during training to ensure that inverse mappings are accurate as well.
In addition, we also instrument a {\em Hodrick-Prescott filter} in post-processing to mitigate false positives. 
We evaluate TSI-GAN using 250 well-curated and harder-than-usual datasets and compare with 8 state-of-the-art baseline methods. The results demonstrate the superiority of TSI-GAN to all the baselines, 
offering an overall performance improvement of 13\% and 31\% over the second-best performer MERLIN 
and the third-best performer LSTM-AE, 
respectively.
\vspace{-5pt}
\vspace{-10pt}
\end{abstract}
\section{Introduction}
\vspace{-10pt}
Anomaly detection aims to identify sub-sequences of various lengths that are considered abnormal within a context represented by data. Accurate and automated anomaly detection is crucial to a wide range of applications including network security, smart manufacturing, autonomous driving, and digital healthcare. Time-series data is ubiquitous in almost all application domains; hence, time-series anomaly detection has been actively studied for years, especially recently using machine learning. However, it remains a very challenging task for three key reasons: (i) labels for abnormal data are often rare, preventing proper training of supervised learning models; (ii) real-world time-series data is often subject to noise and characterized by complex temporal patterns that are difficult to identify; (iii) different datasets have different properties and thus often require a specific choice of parameters (e.g., using domain knowledge) for anomaly detectors to work well, making them hard to generalize.

To address these challenges, we propose a novel generative adversarial network (GAN) architecture called TSI-GAN for {\em unsupervised} time series anomaly detection. First, we encode the input time series to images to capture the temporal correlation and various types of deviance present in the time series, which explains part of our approach, TSI, which stands for {\em Time Series to Images}. This encoding also allows us to leverage GAN's outstanding performance on tasks of image generation \cite{NIPS2017_ImproveWassersteinGAN} and image-to-image translation \cite{isola2017image}. Second, we design a GAN with two critics and two generators that consist of convolutional layers in order to reconstruct the encoded images and obtain effective reconstruction errors. The purpose of the GAN is to learn a generalized distribution of normal samples such that it produces reconstruction errors that are (i) large on anomalous inputs and (ii) small on normal data even in the presence of noise and time non-stationarity. We also take a fully nonparametric approach throughout our design pipeline and as a result our model does not make any assumptions about the underlying data and does not require choosing parameters for each dataset, or altering model architectures like \cite{zenati2018adversarially}.

In addition, GAN-based methods typically sample a random latent and optimize it using gradient descent as a separate step during {\em inference} to find the latent representation that would yield an accurate inverse mapping for {\em each sample} \cite{schlegl2017unsupervised,li2019mad}. This is highly inefficient on large datasets and impractical for real-time applications as proven by \cite{zenati2018adversarially}. In contrast, we train an encoder-decoder network in our GAN with {\em cycle consistency loss} to obtain the latent representation of the inverse mapping automatically and immediately, making our inference almost instantaneous.

Third, as a further enhancement we address false positives (alarms), which are often a pain point in existing anomaly detection methods. To this end, we post-process the reconstruction errors using the {\em Hodrick-Prescott filter} \cite{hodrick1997postwar} and then combine the errors from two encoding channels using a weighted sum. This way, we obtain a reliable anomaly score vector which leads to reduced false positives.


In summary, this paper makes the following contributions:
\begin{itemize}[leftmargin=1em]
    \item We introduce TSI-GAN, a novel convolutional cycle-consistent GAN architecture that learns to reconstruct 2D-encoded complex 1D time-series data and produces reliable reconstruction errors for detecting non-trivial time series anomalies without any labels, and in real-time. 
    
    \item We address the challenge of model generalization by taking a fully nonparametric approach throughout our design pipeline. As a result, our method makes no assumptions about underlying data and requires no manual parameter choice, or changing model architectures.

    \item We mitigate false alarms as a common issue in anomaly detection, by post-processing the reconstruction error using a filtering technique and a weighting strategy.

    \item We benchmark TSI-GAN against eight state-of-the-art baseline methods on 250 well-curated and harder-than-usual datasets. The results validate our approach as the best performer overall, with a large winning margin over other methods.
\end{itemize}

Our results are fully reproducible, with code open-sourced at {\tt\footnotesize \url{https://github.com/LabSAINT/TSI-GAN}}.
\vspace{-12pt}

\section{Related work}
\vspace{-5pt}
Due to the importance of anomaly detection in many applications,
research in this field has been active for years. While
statistical methods are traditionally applied, machine learning and especially
deep learning-based approaches have recently received
increasingly more attention due to their attractive performance. These methods can generally be classified into:

{\bf Proximity-based methods} classify a data point as a point anomaly or a sub-sequence as a collective anomaly when its locality is sparsely populated. These methods can be further classified into {\em cluster-based methods} such as k-means clustering \cite{chawla2013k}, {\em distance-based methods} such as k-nearest neighbors \cite{angiulli2002fast}, and {\em density-based methods} such as DBSCAN \cite{ccelik2011anomaly}. The main drawback of these methods when applied to time series anomaly detection is that they require the number of anomalies to be known a priori and are unable to capture temporal patterns. Time-series discord discovery \cite{yankov2008disk} is a recently proposed distance-based method that identifies very unusual subsequences in a time series. Under this category, Nakamura et al. introduced MERLIN \cite{nakamura2020merlin}, which is considered to be the state-of-the-art for anomaly detection in univariate time series and is included as a baseline in our experiments.

{\bf Prediction-based methods} try to predict future values of a time series and classify a data point as an anomaly if the predicted value differs from the real data by more than a specified threshold. Time series forecasting methods such as ARIMA \cite{pena2013anomaly} can be used, but they often require extensive examination and preprocessing of data and are sensitive to parameters. Several deep-learning approaches have been proposed to overcome these limitations. For example, Hundman et al. \cite{hundman2018detecting} proposed an LSTM model with dynamic thresholding (LSTM-DT) to make predictions and reduce false positives.

{\bf Reconstruction-based methods} learn a latent low-dimensional representation of the input time-series data and try to reconstruct the input based on the representation. The assumption is that anomalies will lose information when mapped to the latent space and thus will not be reconstructed accurately, producing a larger reconstruction error. Hence, reconstruction error is measured at each time step and thresholding techniques are applied to detect the anomalies. Several deep learning approaches have been proposed including LSTM-based Autoencoder (LSTM-AE) \cite{nguyen2021forecasting}, Dense Autoencoder (DENSE-AE) \cite{sakurada2014anomaly}, DONUT \cite{donut} which uses a Variational Autoencoder (VAE), and GAN-based methods \cite{geiger2020tadgan,schlegl2017unsupervised,li2019mad}. TadGAN~\cite{geiger2020tadgan} presents a recent study using GAN to perform this task and it is considered to be state-of-the-art in terms of GAN-based methods.

However, TadGAN uses 1-D representation and requires the sampling interval of input data to be known for data preprocessing; otherwise, anomalies that do not have extreme amplitude (either high or low relative to other points) will not be detected. This is a notable limitation because most anomalies in the real world are complicated rather than just simple amplitude spikes or dips. Another related work is T2IVAE~\cite{xu2021nvae}, which transforms time series to images 
and uses VAE to reconstruct the 
input time series. However, VAEs are prone to overfitting and often reconstruct anomalous samples quite accurately, resulting in unreliable reconstruction errors. Even though T2IVAE attempts to reduce this risk by employing an adversarial training strategy in the last five training epochs, the overfitting effect remains rather prominent.


We take a GAN-based approach instead of VAE because we find that GAN is strongly averse to the overfitting phenomenon when it comes to infrequent anomalous samples and unlike TadGAN we 2D encode the input time series and use CNN layers in our GAN to learn feature maps as if learning from images. This way, we are able to encode temporal information/correlation and capture various types of deviance and thus obtain more accurate and reliable anomaly scores based on reconstruction errors.

There are also {\bf commercial tools} including Microsoft Azure Anomaly Detector \cite{ren2019time} and LinkedIn Luminol \cite{luminol_code}. Azure uses spectral residual (SR) from the saliency detection domain \cite{ren2019time} and CNN to learn a discriminating threshold. The output is a sequence of labels indicating if a particular timestamp is anomalous. Luminol uses the Bitmap detector algorithm \cite{wei2005assumption} which divides input time series into chunks and calculates the frequency of similar chunks to calculate anomaly scores. These commercial tools are included as baselines in our experiments as well.

\section{Encoding Time-series to Images}\label{section:encoding}
\vspace{-8pt}

\begin{figure}[t]
    \centering
    \subfloat[Normal window]{
        \includegraphics[width=0.28\columnwidth,
        ]{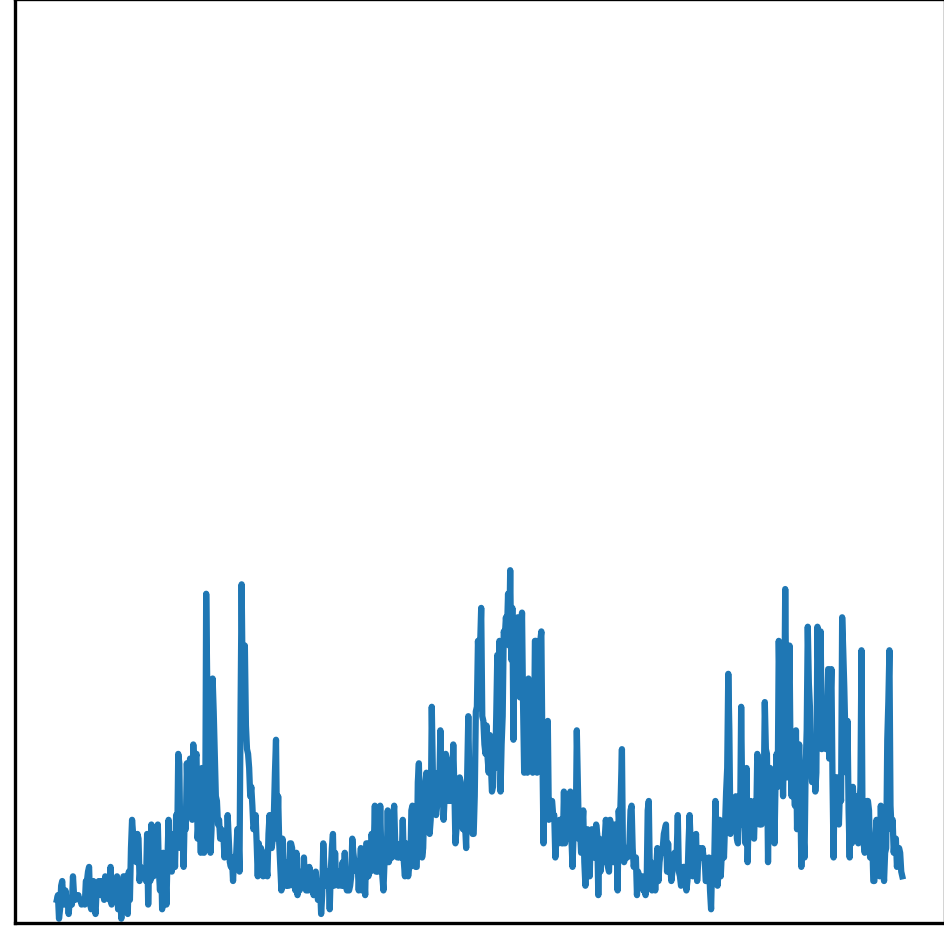}%
        \label{fig:normal_window}%
    } 
    \subfloat[GAF representation]{
        \includegraphics[width=0.28\columnwidth,
        ]{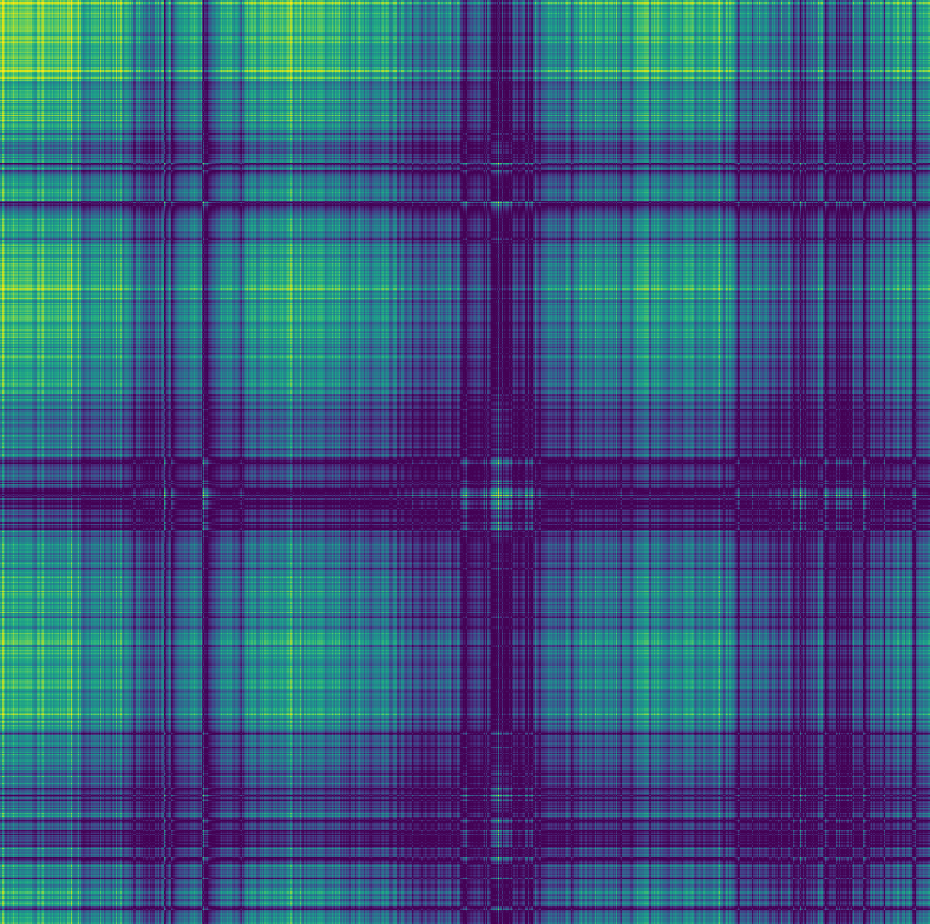}%
        \label{fig:normal_gaf}%
    }
    \subfloat[RP representation]{
        \includegraphics[width=0.28\columnwidth,
        ]{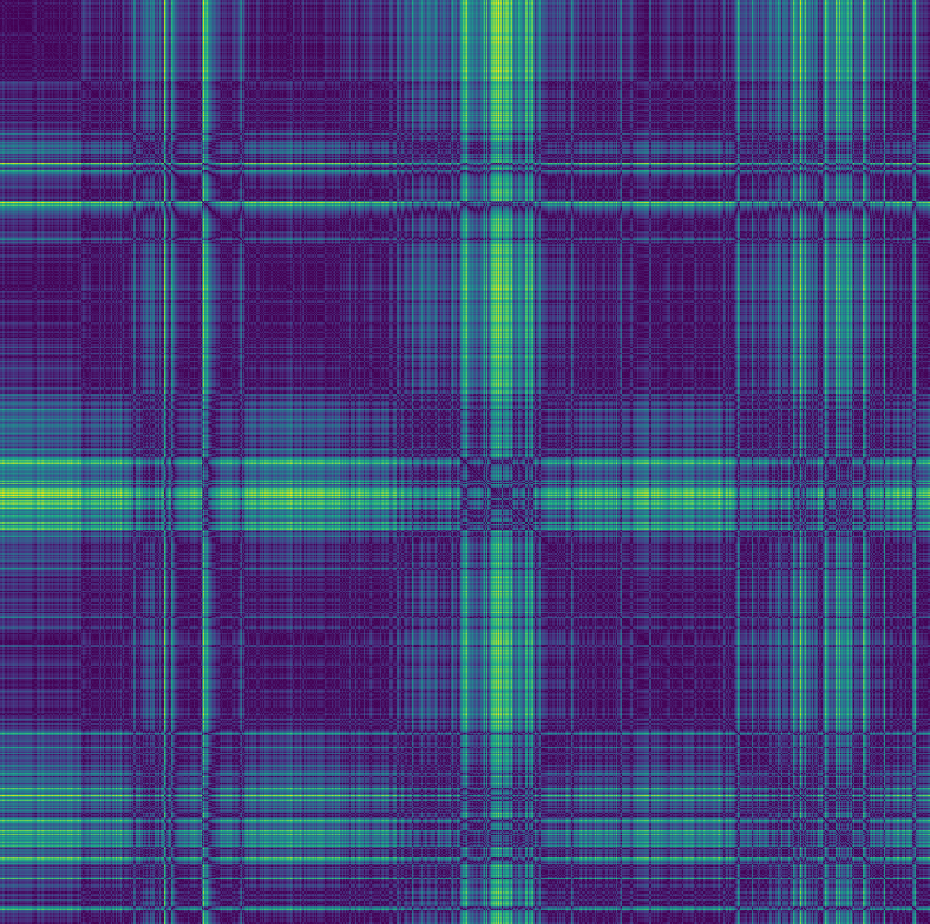}%
        \label{fig:normal_rp}%
    }
    
    \subfloat[Anomalous window]{
        \includegraphics[width=0.28\columnwidth]{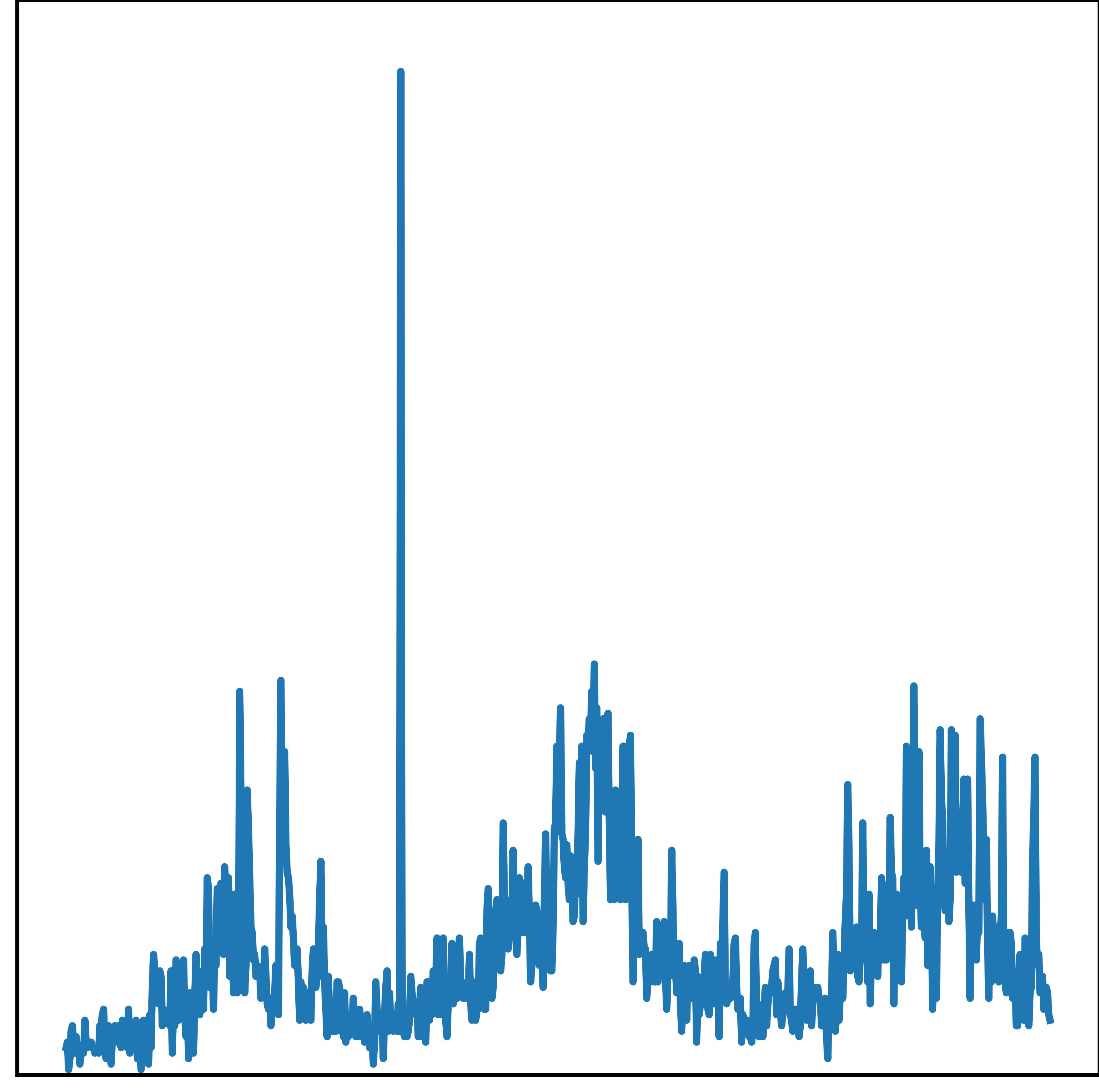}%
        \label{fig:abnormal_window}%
    } 
    \subfloat[GAF representation]{
        \includegraphics[width=0.28\columnwidth,]{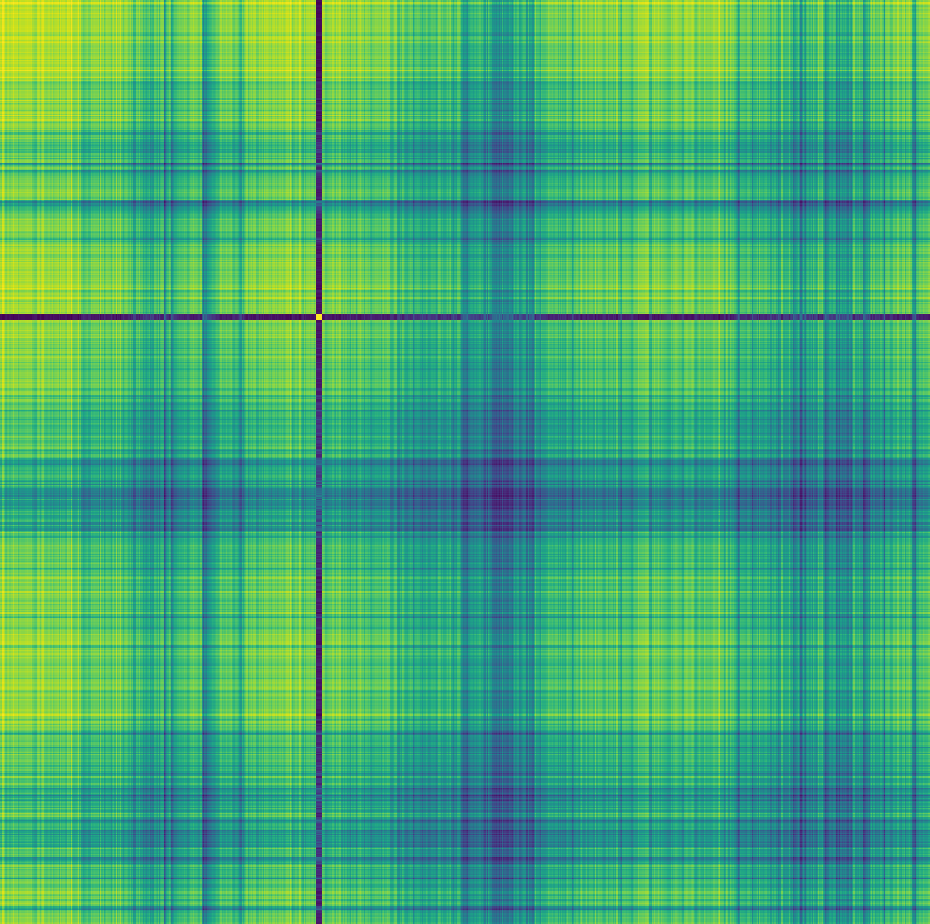}%
        \label{fig:abnormal_gaf}%
    }
    \subfloat[RP representation]{
        \includegraphics[width=0.28\columnwidth]{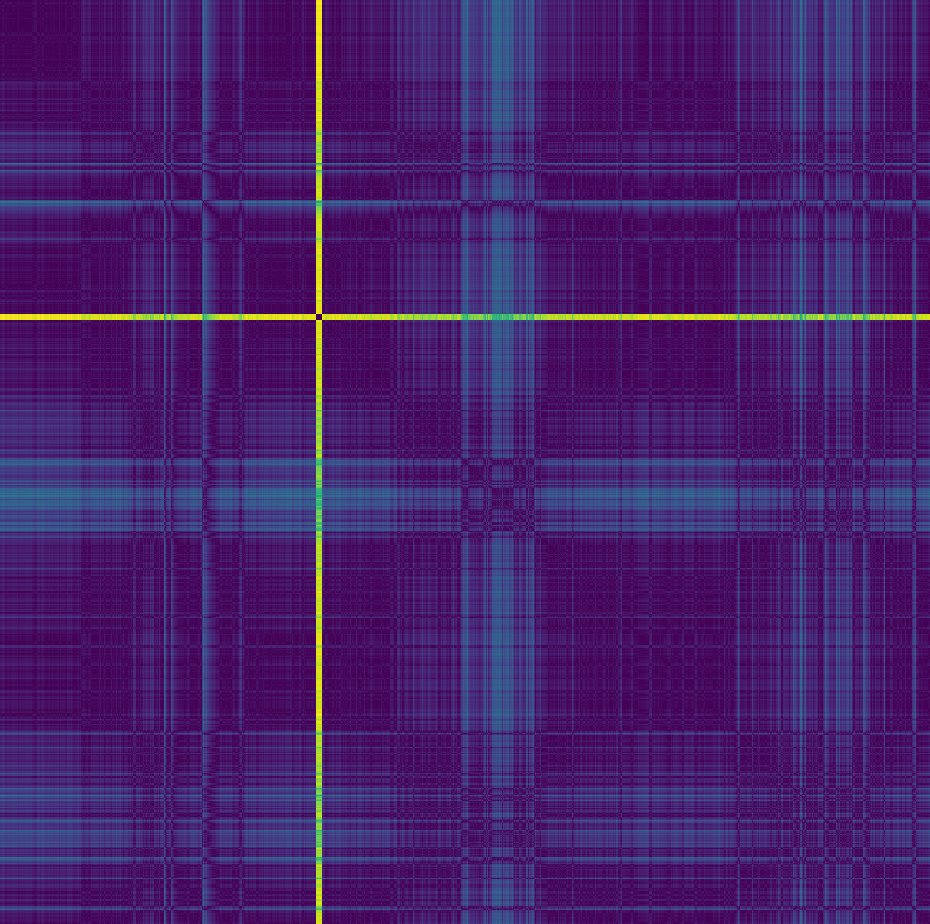}%
        \label{fig:abnormal_rp}%
    }%
    \caption{Illustration of normal (top row) and anomalous (bottom row) windows encoded by GAF and RP, for an example time-series.}
    \label{fig:example_windows}
    \vspace{-16pt}
\end{figure}

The core idea behind encoding the time-series to images is that if any time step is anomalous, then the row and column corresponding to that time step in the encoded image will be significantly different from other normal pixels (see \fref{fig:example_windows}) and thus could be easily detected by a reconstruction-based model. Consider an input time series $\mathring{X}=\{x_1, x_2,...,x_T\}$, where $T$ is the time series length. We use a sliding window with window size $W$ and step size $S$ to divide $\mathring{X}$ into $N$ overlapping sub-sequences, 
$\mathring{X}_k=\{x_{k+1}, x_{k+2}, \dots , x_{k+W}\}$, where $k=0, \dots, N-1$ and $N = \lfloor \frac{T - W}{S} \rfloor$. We set $W=64$ and $S=1$ and convert each window of size $64$ into a two-channel image of size $64 \times 64 \times 2$, using two time-series encoding techniques: Gramian Angular Field (GAF)~\cite{wang2015encoding} and Recurrence Plot (RP)~\cite{eckmann1995recurrence}.
\vspace{-14pt}
\subsection{Gramian Angular Field (GAF)}
\label{sec:gaf}
\vspace{-4pt}
Given a sub-sequence $\mathring{X}_k=\{x_{k+i}\}_{i=1}^W$ at time step $k$, GAF rescales all the observations into the interval $[-1,1]$ and calculates $\bar{X}_k = \{\bar{x}_{k+i}\}_{i=1}^W$, where
\vspace{-2pt}
\begin{equation*}\label{eq:GAF_element}
\vspace{-2pt}
    \bar{x}_{k+i} = \frac{(x_{k+i} - \max(\mathring{X}_k)) + (x_{k+i} - \min(\mathring{X}_k))} {\max(\mathring{X}_k) - \min(\mathring{X}_k)}
\vspace{-2pt}
\end{equation*}
Next, we represent each rescaled $\bar{X}_k$ using polar coordinates, as radius $r=t_{k+i}/W$ where $t_{k+i} \in \mathbb{N}$ is the timestamp, and angular  $\phi=\arccos(\bar{x}_{k+i}) \in [0, \pi]$. 
This polar conversion produces a one-to-one mapping with a unique inverse function and preserves absolute temporal relation (as opposed to Cartesian coordinates).
Thus, we can identify the temporal correlation at different time intervals by calculating the trigonometric sum between each point within the sub-sequence:
\vspace{-8pt}
\begin{equation*}
X_{k}^{GAF} = 
(\bar{X}_k)^T \otimes \bar{X}_k - \nonumber 
\left(\sqrt{I - (\bar{X}_k)^2}\right)^T \otimes \sqrt{I -(\bar{X}_k)^2}
\vspace{-6pt}
\end{equation*}
where $X_{k}^{GAF}$ is a $W\times W$ matrix, $I$ is the unit row vector $[1,1,...,1]$ ($\bar{X}_k$ is a row vector too), and $\otimes$ represents outer product. 

\vspace{-15pt}
\subsection{Recurrence Plot (RP)}\label{sec:rp}
\vspace{-4pt}
A recurrence plot (RP) \cite{eckmann1995recurrence} is an image that represents the distance between observations 
extracted from a sub-sequence time series. 
Given a sub-sequence window $\mathring{X}_k=\{x_{k+i}\}_{i=1}^W$, we calculate a RP matrix $X_{k}^{RP}$ of dimension $W\times W$ where each element at row $a$ and column $b$ is defined as
\begin{align}
    \vspace{-6pt}
    x_{k,(a,b)}^{RP}=\Theta( \epsilon - \|x_{k+a} - x_{k+b}\|),
    \; \forall a,b\in\{1,\cdots,W\}
    \vspace{-6pt}
\end{align}
where $\Theta(\cdot): \mathbb{R} \rightarrow \{0,1\}$ is a Heaviside function, and $\epsilon$ is a predefined distance threshold. 
In this work, instead of using binary representation, we use raw distances $\|x_{k+a} - x_{k+b}\|$ (without the need for choosing $\epsilon$ or $\Theta(\cdot)$) to construct the RP matrix; the resulting 2D image will thus have more granularity scales of the distances.
In order to align RP images with GAF images on the same scale, we scale the RP matrix into the range $[-1,1]$ before further processing.

\vspace{-12pt}
\subsection{Combining two channels}
\vspace{-4pt}
After encoding the series using GAF and RP, respectively, we treat them as two channels and stack them along the channel axis to obtain:
\vspace{-2pt}
\begin{equation*}
    X_k = Stack(X_{k}^{GAF},X_{k}^{RP})
\vspace{-2pt}
\end{equation*}
Since we have divided the original time series $\mathring{X}$ into $N$ overlapping sub-sequence windows of size $W$, and encoded each window as a 2-channel image of shape $[W\times W \times 2]$, we thus finally obtain a sequence of images $\mathbb{X}=\{{X}_{k}\}_{k=1}^N$.
\vspace{-10pt}
\section{The TSI-GAN Model}
\vspace{-8pt}
\subsection{Model Architecture}
\vspace{-8pt}
Reconstruction-based anomaly detection methods learn a model that maps input data (in our case, an image with two channels) to the latent low-dimensional space and then reconstructs the input using the latent representation. The objective is to train a model that captures a generalized latent representation of the {\em normal} patterns, such that anomalies will {\em not} be reconstructed accurately and hence result in a larger reconstruction error. In our proposed method, we learn two mapping functions, $\mathcal{E} : \mathcal{X} \to \mathcal{Z}$ and $\mathcal{G} : \mathcal{Z} \to \mathcal{X}$, where $\mathcal{X}$ represents the input domain, 
$\mathcal{Z}$ represents the latent domain for which Gaussian distribution $\mathcal{N}(0,1)$ is used. For any given input image at time step $k$, denoted by $X_k$, the model tries to reconstruct it as $X_k \to \mathcal{E}(X_k) \to \mathcal{G}(\mathcal{E}(X_k)) \approx \hat{X}_k$.

The entire model architecture is presented in Fig.~\ref{fig:modelArchitecture}. We model the above mapping functions as Generators, where $\mathcal{E}$ acts as an encoder which maps the input image to the latent space using convolution layers, and $\mathcal{G}$ acts as a decoder which transforms the latent representation to a reconstructed input image using transposed convolution. We use two Critics $\mathcal{C}_x$ and $\mathcal{C}_z$:
$\mathcal{C}_x$ regulates the decoder $\mathcal{G}$ by trying to distinguish real images $X$ from the reconstructed images $\mathcal{G}(\mathcal{E}(x))$; $\mathcal{C}_z$ regulates the encoder $\mathcal{E}$ by trying to liken the latent representation $\mathcal{E}(x)$ to the Gaussian noise $z$. The $L_2$-norm will be used in our cycle consistency loss which we describe later in \sref{sec:ccl}.

\vspace{-14pt}
\subsection{Loss Function and Training Strategy}
\vspace{-8pt}
We use two loss functions: (1) Wasserstein loss, to match the distribution of generated images with the distribution  of input images, and (2) cycle consistency loss, to ensure the desired mapping route $X_k \to Z_k \to \hat{X}_k$.
\vspace{-14pt}
\subsubsection{Wasserstein Loss:}
\label{subsec:WassersteinLoss}

We train the generator $\mathcal{G}$ and its critic $C_x$ with Wasserstein loss:
\thinmuskip=1mu
\medmuskip=2mu plus 1mu minus 2mu
\thickmuskip=3mu plus 2mu minus 3mu
\begin{equation}\label{objective:1}
\min_{\mathcal{G}}\max_{\mathcal{C}_x \in \mathbf{C}_x} 
L_X(\mathcal{C}_x,\mathcal{G}) \triangleq
\mathbb{E}_{x\sim \mathbb{P}_{X}}[\mathcal{C}_x(x)] - \mathbb{E}_{z\sim \mathbb{P}_{Z}}[ {\mathcal{C}_x(\mathcal{G}(z))}]
\end{equation}
\thinmuskip=3mu
\medmuskip=4mu plus 2mu minus 4mu
\thickmuskip=5mu plus 5mu

Similarly, for Encoder $\mathcal{E}$ and its Critic $\mathcal{C}_z$, the loss function is defined as:
\thinmuskip=1mu
\medmuskip=2mu plus 1mu minus 2mu
\thickmuskip=3mu plus 2mu minus 3mu
\begin{equation} \label{objective:2}
\min_{\mathcal{E}}\max_{\mathcal{C}_z \in \mathbf{C}_z} L_Z(\mathcal{C}_z,\mathcal{E}) \triangleq
\mathbb{E}_{z\sim \mathbb{P}_{Z}}[\mathcal{C}_z(z)] -  \mathbb{E}_{x\sim \mathbb{P}_{X}}[ {\mathcal{C}_z(\mathcal{E}(x))}]
\end{equation}
\thinmuskip=3mu
\medmuskip=4mu plus 2mu minus 4mu
\thickmuskip=5mu plus 5mu
where $\mathbf{C}_x$ and $\mathbf{C}_z$ are the set of all the 1-Lipschitz functions.

\begin{wrapfigure}{R}{0.59\columnwidth}
\centering
\vspace{-20pt}
\resizebox{0.59\columnwidth}{!}{
\includegraphics{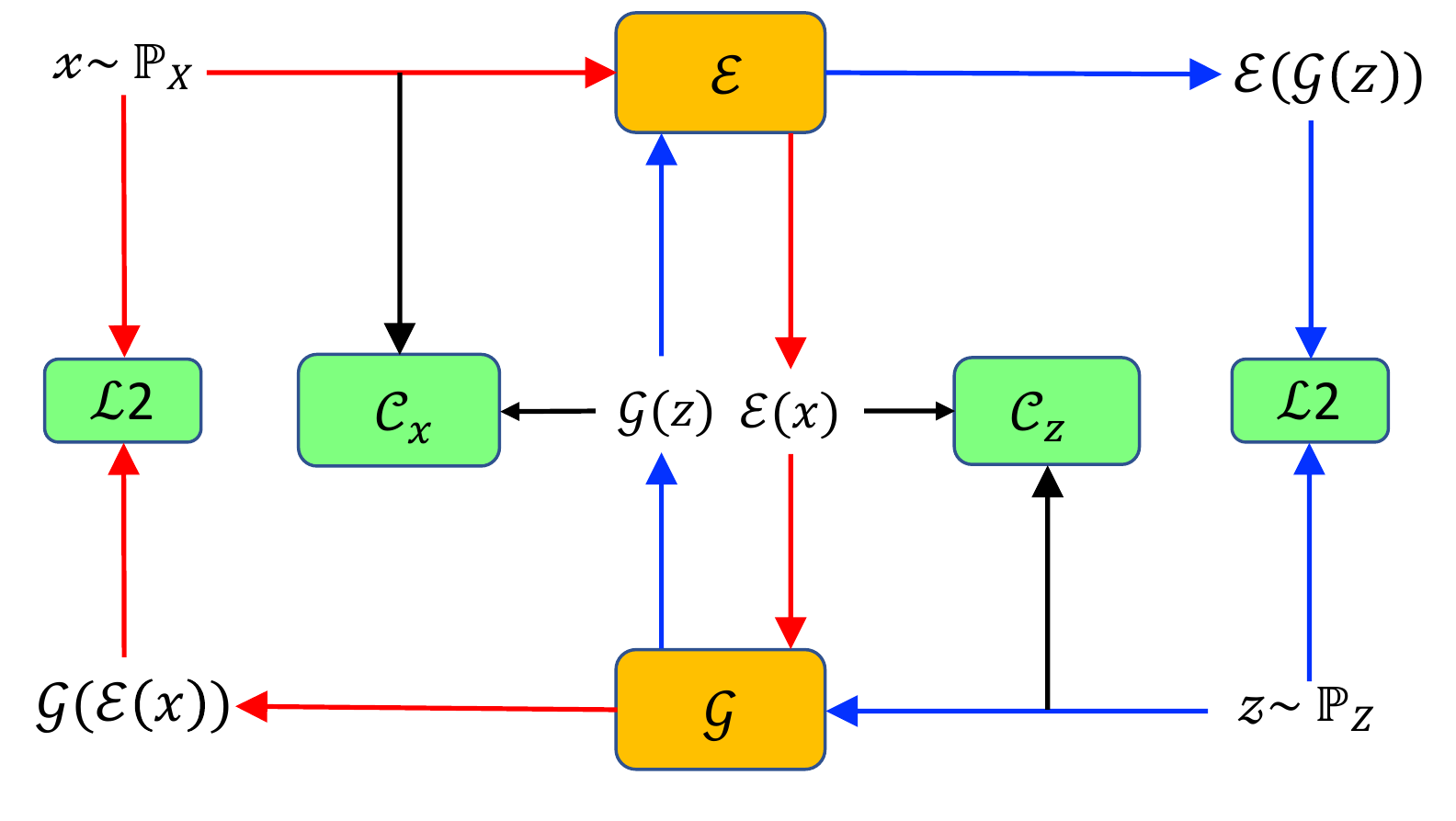}
}
\caption{TSI-GAN model architecture: 
$\mathcal{E}$ is the encoder and $\mathcal{G}$ the decoder; $\mathcal{C}_x$ and $\mathcal{C}_z$ are the critics.}
\label{fig:modelArchitecture}
\vspace{-20pt}
\end{wrapfigure}

We also add a gradient penalty regularization term to both (\ref{objective:1}) and (\ref{objective:2}) to ensure a 1-Lipschitz continuous Critic so that Wasserstein Loss validly approximates the {\em Earth Mover's Distance} \cite{levina2001earth}. The complete architecture is presented in Table \ref{tab:architecture} and Figure~\ref{fig:modelArchitecture}.

\vspace{-14pt}
\subsubsection{Cycle Consistency Loss:}\label{sec:ccl}

The GAN model described above is able to map $X_k$ to a desired $Z_k$. However, the inverse mapping of $Z_k$ back to $\hat{X}_k$ is not guaranteed by training with just Wasserstein losses alone. This is because those losses only ensures distribution similarity but not instance similarity. To this end, Schlegl et al. \cite{schlegl2017unsupervised} proposed an iterative approach where they sample a random latent and optimize it using gradient descent as a separate step during {\em inference} to find the best $Z_k$ that would generate $\mathcal{G}(\mathcal{E}(X_k))$ that is most similar to the input image $X_k$. However, this method suffers from large search space and is inefficient for large datasets and real-time applications, as shown by Zenati et
al. \cite{zenati2018adversarially}. Hence, we use {\em cycle-consistency loss} \cite{zhu2017unpaired} to train the generators $\mathcal{E}$ and $\mathcal{G}$:
\vspace{-5pt}
\thinmuskip=3mu
\medmuskip=4mu plus 2mu minus 4mu
\thickmuskip=5mu plus 5mu
\begin{equation} \label{objective:3}
    \min_{\mathcal{E}} L_{CL}(\mathcal{E}) \triangleq  \mathbb{E}_{x\sim \mathbb{P}_{X}}\| x - \mathcal{G}(\mathcal{E}(x)) \|_{2} + \mathbb{E}_{z\sim \mathbb{P}_{Z}}\| z - \mathcal{E}(\mathcal{G}(z))
    \|_{2}
\vspace{-10pt}
\end{equation}
For $\mathcal{G}$ we only use the forward consistency loss, as the backward consistency loss (i.e., $\mathbb{E}_{z\sim \mathbb{P}_{Z}}\| z - \mathcal{E}(\mathcal{G}(z))
    \|_{2}$) has been integrated into \eqref{objective:3} and thus is not necessary for $\mathcal{G}$: 
\vspace{-5pt}
\thinmuskip=3mu
\medmuskip=4mu plus 2mu minus 4mu
\thickmuskip=5mu plus 5mu
\begin{equation} \label{objective:4}
    \min_{\mathcal{G}} L_{CL}(\mathcal{G}) \triangleq  \mathbb{E}_{x\sim \mathbb{P}_{X}}\| x - \mathcal{G}(\mathcal{E}(x)) \|_{2}.
\vspace{-15pt}
\end{equation}
\subsubsection{Final Objective}

Combining the objectives (\ref{objective:1}), (\ref{objective:2}), (\ref{objective:3}), (\ref{objective:4})  we arrive at the final objective:
\vspace{-10pt}
\begin{equation}
    \min_{\{\mathcal{E},\mathcal{G}\}}\max_{\{\mathcal{C}_x \in \mathbf{C}_x, \mathcal{C}_z \in \mathbf{C}_z\}} L_X(\mathcal{C}_x,\mathcal{G}) + 
    L_Z(\mathcal{C}_z,\mathcal{E}) + 
    L_{CL}(\mathcal{E}) +
    L_{CL}(\mathcal{G})
\end{equation}
\vspace{-30pt}


\subsection{Post-processing and Anomaly Detection}
\vspace{-5pt}
Unlike other anomaly detection methods, we added a post-processing procedure to achieve a more reliable detector, as follows. 
We then extract the reconstruction error for each channel as $\epsilon_{gaf}$ and $\epsilon_{rp}$ (refer lines 1-5 in Algorithm \ref{alg:detecting}). Calculating thresholds directly on the raw reconstruction error will lead to many false positives. To mitigate this, we smooth the reconstruction error to suppress frequently occurring minor error peaks which are usually caused by normal behavior rather than anomalies. We use the {\em Hodrick-Prescott filter} \cite{hodrick1997postwar} because of its excellent capability of removing short-term fluctuations in data since we are only concerned with peaks that persist for a sustained period of time. It extracts a smooth trend $r$ from a given sequence $\epsilon$ of length $N$ by solving:
\vspace{-6pt}
\thinmuskip=1mu
\medmuskip=0mu minus 2mu
\thickmuskip=0mu minus 3mu
\begin{equation*}
    \min_{r} \left( \sum_{k=1}^{N}(\epsilon_k - r_k)^2 + \lambda\sum_{k=2}^{N-1}[(r_{k+1} - r_{k}) - (r_k - r_{k-1})]^2 \right)
\end{equation*}
\vspace{-40pt}

\begin{wrapfigure}{L}{0.61\columnwidth}
\vspace{-24pt}
\begin{minipage}{0.61\columnwidth}
\begin{algorithm}[H]
\DontPrintSemicolon
\caption{Anomaly Detection using TSI-GAN}\label{alg:detecting}
\tcp{Compute reconstruction errors}
\For{$k = 1,...,N$} {
$\hat{X}_k \gets \mathcal{G}(\mathcal{E}(X_k))$\;
$\hat{X}_k^{GAF},\hat{X}_k^{RP} \gets$ {\small extract GAF \& RP channels from} $\hat{X}_k$\;
$\epsilon_{gaf_k} \gets  \sum_{i=1}^{W}\sum_{j=1}^{W}({X}_{k,i,j}^{GAF} - \hat{X}_{k,i,j}^{GAF})^2 $\;
$\epsilon_{rp_k} \gets \sum_{i=1}^{W}\sum_{j=1}^{W}({X}_{k,i,j}^{RP} - \hat{X}_{k,i,j}^{RP})^2$\;
}
\tcp{Post-processing to obtain anomaly scores}
\For{$ch \in \{gaf, rp\}$} {
$\epsilon_{ch} \gets$ HP$(\epsilon_{ch})$ \tcp{Hodrick-Prescott filter} 
$peaks_{ch} \gets$ find\_peaks$(\epsilon_{ch})$\; 
$\sigma_{ch} \gets \frac{(peaks_{ch}[0] - peaks_{ch}[1])}{peaks_{ch}[0]} + 1$\; 
}

$score\_vec \gets \sigma_{gaf} \times \epsilon_{gaf} + 
\sigma_{rp} \times \epsilon_{rp}$\; \label{alg:score}
\tcp{Detect anomalies}
$mean \gets$ mean($score\_vec$)\;
\For{$k = 1,...,N$} {
\uIf{$score\_vec_k > mean $} {
$pred_k = true$\;
}
\Else {
$pred_k = false$\;
}
}
Group consecutive $pred_k$'s into $\{seq_i\}_{i=1}^L$ \;
\tcp{Pruning to reduce false alarms}
$\{m_i\} \gets {\max(\{seq_i\})}$ \;
$\{m_i\} \gets$ sort$(\{m_i\}, descending=true)$\;
sort $\{seq_i\}$ in the same order of $\{m_i\}$\;
\For{$i=1,...,L$} {
$p_i \gets (m_{i-1} - m_{i})/m_{i-1}$\;
\If{$p_i < \theta$} {
reclassify $\{seq_j\}_{j=i}^L$ as normal\;
break;
}
}
\end{algorithm}
\end{minipage}
\vspace{-29pt}
\end{wrapfigure}

\vspace{30pt}
After smoothing $\epsilon_{gaf}$ and $\epsilon_{rp}$, 
we find the local (neighborhood) peaks in each channel and sort them in descending order to calculate a confidence level $\sigma  \in  [1,2]$ for each channel:
\thinmuskip=3mu
\medmuskip=4mu plus 2mu minus 4mu
\thickmuskip=5mu plus 5mu
\begin{equation*}
    \sigma =  \frac{peaks[0] - peaks[1]}{peaks[0]} + 1,
\end{equation*}
where $peaks[0]$ and $peaks[1]$ are the first and the second highest peaks in the smoothed reconstruction errors, respectively. The idea is that when the difference between these two peaks is large, that channel is assumed to be more confident about its detection of the anomaly and hence weighed higher in the final anomaly score. This score is defined by combining the two reconstruction errors $\epsilon_{gaf}$ and $\epsilon_{rp}$ using their respective confidence level (see line \ref{alg:score} in Algorithm \ref{alg:detecting}).
Here it is defined as a vector (of length $N$) because each of the $N$ windows will have an anomaly score. For the weight $\sigma$, if there are multiple anomalies, the difference between $peaks[0]$ and $peaks[1]$ will be small and thus $\sigma$ will be smaller than the other channel if the other channel detects a single outlier, which is desired since outliers are rare by definition and thus single outliers are more likely than multiple. Otherwise, if both channels detect multiple, they will be weighted by similar $\sigma$'s.

After obtaining the anomaly score for each window, we calculate the mean anomaly score over all the windows and any window that exceeds this threshold is flagged as an anomaly. Following that, consecutive anomalous windows will be grouped together to form a sequence (i.e., collective anomaly). Finally, we use an anomaly pruning approach (lines 17-25 in Algorithm \ref{alg:detecting}) introduced by Hundman et al.~\cite{hundman2018detecting}, to further mitigate false positives. The above post-processing and detection procedures are formulated in Algorithm~\ref{alg:detecting}.
\vspace{-5pt}

\begin{wraptable}{r}{0.55\columnwidth}
\vspace{-68pt}
\resizebox{0.55\columnwidth}{!} {
\centering
  \begin{tabular}[H]{cccccc}
    \toprule
    Operation&Kernel&Strides&Units&BN?&Activation\\
    \midrule
    \textbf{Encoder} \\
    Convolution&$7\times7$&$3\times3$&$48$&\checkmark&ReLU\\
    Convolution&$5\times5$&$3\times3$&$96$&\checkmark&ReLU\\
    Convolution&$4\times4$&$2\times2$&$192$&\checkmark&ReLU\\
    Convolution&$2\times2$&$1\times1$&z\_dim&$\times$&$-$\\
    
    \midrule
    \textbf{Decoder} \\
    Transp. Conv &$2\times2$&$1\times1$&$192$&\checkmark&LReLU\\
    Transp. Conv &$4\times4$&$2\times2$&$96$&\checkmark&LReLU\\
    Transp. Conv &$5\times5$&$3\times3$&$48$&\checkmark&LReLU\\
    Transp. Conv &$7\times7$&$3\times3$&$2$&$\times$&Tanh\\
    \midrule 
    \textbf{Critic X} \\
    Convolution&$7\times7$&$3\times3$&$48$&LN&LReLU\\
    Convolution&$5\times5$&$3\times3$&$96$&LN&LReLU\\
    Convolution&$4\times4$&$2\times2$&$192$&LN&LReLU\\
    Convolution&$2\times2$&$1\times1$&$1$&$\times$&$-$\\ 
    \midrule
    \textbf{Critic Z} \\ 
    Fully Conn.&&&$50$&LN&ReLU\\
    Fully Conn.&&&$25$&LN&ReLU\\
    Fully Conn.&&&$1$&$\times$&$-$\\
    \midrule
    {\bf Hyperparams.}& & & & &\\
    z\_dim&$100$  & Lrn. rate ($\alpha$) & 1e-4 & Iterations&$5000$\\
    Optimizer&RMSProp & Wt decay ($\lambda_{wd}$) & 1e-4& BatchSize& $128$\\
    
    \bottomrule
  \end{tabular}
  }
    \caption{\footnotesize Architecture of our proposed TSI-GAN. Transp. Conv: Transposed Convolution; BN: Batch Normalization; LN: Layer Normalization respectively; LReLU: LeakyReLU; Lrn. rate: learning rate.}
  \label{tab:architecture}
  \vspace{-20pt}
  \end{wraptable}
\vspace{-16pt}
\section{Performance Evaluation}\label{sec:exp}
\vspace{-12pt}
\subsection{Datasets}\label{sec:datasets}
\vspace{-8pt}

We use the UCR {2021} anomaly detection dataset\footnote{UCR 2021 anomaly detection dataset: \url{https://bit.ly/3V2n6FY}} which contains 250 sub-datasets collected from a variety of sources. Unlike commonly used datasets such as Yahoo, Numenta, and NASA which are found to have numerous flaws \cite{wu2021current} including incorrect ground truth labels, triviality of the anomalies, and unrealistic anomaly density, this UCR dataset is carefully curated, harder to detect and is much more reliable. Moreover, this dataset contains a combination of point, collective, and contextual anomalies as well as amplitude, seasonal, and trend anomalies, which offers a good variety for evaluation.

\begin{table*}[t]
\resizebox{\columnwidth}{!}{
    \centering
    \begin{tabular}{cccccccc}
    \toprule
    \multirow{2}{*}{\textbf{Property}} & &
        \multicolumn{2}{c}{\textbf{Dataset}} \\
        \cmidrule{2-8}
          & AirTemperature&PowerDemand&InternalBleeding&EPG&NASA T-1&Noise&All datasets\\
    \midrule
    \# Sub-datasets & 14&8&26&12&10&16&250 \\
    \# Data Points &98208&239448&194992&359304&113488&629494&19353766 \\
    \# Anomalous Points &398&1688&3018&1292&644&3134&49363\\
     \hspace{5pt} \# (\% tot.)&0.004\%&0.007\%&0.015\%&0.003\%&0.005\%&0.004\%&0.002\% \\
    \bottomrule
    \end{tabular}}
    \caption{Statistics of Datasets used in our experiments.}
    \label{tab:dataset}
\vspace{-30pt}
\end{table*}

We choose a total of 6 categories from this dataset and each category contains 4-13 original sub-datasets; each original sub-dataset comes with a distorted duplicate by adding artificial fluctuations. Therefore, the number of sub-datasets is doubled. The only exception is the {\tt Noise} category in which the sub-datasets are chosen from multiple other categories with Gaussian noise added. A brief description of each category is as follows: {\tt AirTemperature} consists of hourly air temperature between 03/01 and 03/31 from 2009 to 2019, collected from CIMIS station 44 in Riverside, CA. {\tt PowerDemand} consists of Italian power demand data between 1/1/1995 and 5/31/1998. {\tt InternalBleeding} consists of the arterial blood pressure measurements of pigs. {\tt EPG} is collected from an insect known as Asian Citrus Psyllid, recorded using an Electropalatography (EPG) apparatus. {\tt NASA T-1} is collected from NASA Mars Science Laboratory (MSL) dataset that consists of spacecraft telemetry signals. Detailed statistics of each category and all the datasets is presented in Table \ref{tab:dataset}.





\begin{table*}[t]
    \resizebox{\columnwidth}{!}{
    \centering
    \begin{tabular}{|c|c|c|c|c|c|c|c|c|c|c|c|c|c|}
    \hline
    \multirow{2}{*}{\textbf{Model}} &
        \multicolumn{2}{c|}{AirTemperature} & \multicolumn{2}{c|}{PowerDemand} & \multicolumn{2}{c|}{InternalBleeding} & \multicolumn{2}{c|}{EPG} &
        \multicolumn{2}{c|}{NASA T-1} & Noise  & \multicolumn{2}{c|}{All 250 datasets}             \\
        \cline{2-11}
        \cline{13-14}
        & Orig. & Distor. & Orig. & Distor. & Orig. & Distor. & Orig. & Distor. & Orig. & Distor. &  & F1 & Precision\\
        \hline
        \rowcolor{green!20}
        TSI-GAN &\textbf{1.0}&\textbf{0.833}&\textbf{0.667}&\textbf{0.667
        }&0.846&0.474&\textbf{0.5}&\textbf{0.556}&\textbf{0.933}&0.267&0.479 & \textbf{0.468} & \textbf{0.445}\\
        \hline
        MERLIN &0.054&0.18&0.04&0.071&\textbf{0.926}&\textbf{0.721}&0.354&0.191&0.613&\textbf{0.6}&\textbf{0.49}&0.414&0.402 \\
        \hline
        LSTM-AE &0.389&0.611&0.375&0.583&0.654&0.308&0.222&0.444&0.533&0.333&0.208&0.355&0.301 \\
        \hline
        DONUT &0.611&0.444&0.083&0.1&0.59&0.564&0.278&0.167&0.333&0.533&0.458&0.351&0.325 \\
        \hline
        LSTM-DT &0.778&\textbf{0.833}&0.25&0.5&0.615&0.449&0.222&0.222&0.6&\textbf{0.6}&0.271&0.32&0.289 \\
        \hline
        DENSE-AE &0.194&0.111&0.0&0.0&0.231&0.077&0.222&0.222&0.2&0.0&0.271&0.159&0.136 \\
        \hline
        TadGAN &0.0&0.133&0.0&0.0&0.282&0.24&0.233&0.189&0.267&0.2&0.171&0.131&0.092 \\
        \hline
        Azure &0.181&0.199&0.083&0.196&0.099&0.176&0.167&0.167&0.007&0.017&0.084&0.05&0.037 \\
        \hline
        Luminol &0.022&0.021&0.078&0.089&0.118&0.046&0.037&0.088&0.009&0.014&0.019&0.049&0.021 \\
        \hline
        
    \end{tabular}}
    \caption{Average F1-Score on original and distorted datasets for each category, as well as F1-Score and Precision averaged over all the 250 datasets.}
    \label{tab:results}
    \vspace{-30pt}
\end{table*}

\vspace{-15pt}
\subsection{Performance Metrics}
\vspace{-9pt}
In real-world application scenarios, most anomalies happen in the form of collective anomalies and hence we use the window-based rules introduced by Hundman et al. \cite{hundman2018detecting}: 
(1) If an anomalous window overlaps any predicted window, a true positive (TP) is recorded; (2) If a predicted window does not overlap with any anomalous window, a false positive (FP) is recorded; (3) If an anomalous window does not overlap with any predicted window, a false negative (FN) is recorded. Based on this set of rules, we calculate {\it Precision} and {\it F1-Score} as the performance metrics.
\vspace{-15pt}
\subsection{Experimental Results}
\vspace{-8pt}
Table \ref{tab:results} reports the average F1-Score on the original and distorted datasets for each category, and in the last column, the F1-Score and Precision averaged over all the 250 datasets.

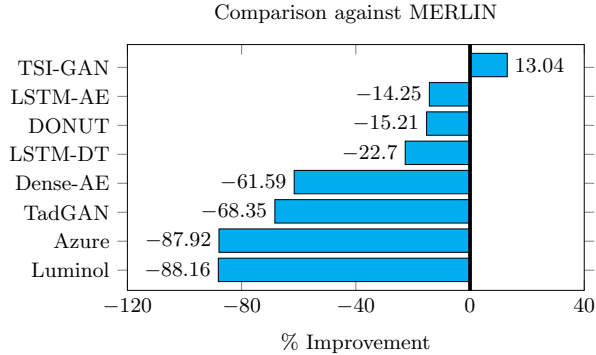
\begin{wrapfigure}{R}{0.6\columnwidth}
\vspace{-35pt}
\centering
\small
\resizebox{0.6\columnwidth}{!} {
\begin{tikzpicture}
\begin{axis} [title = Comparison against MERLIN, xbar,
    bar width = 10pt,
    xmin = -120,
    xmax = 40,
    xtick distance = 40,
    ytick distance = 0.99,
    y=12.5,
    symbolic y coords = {0,Luminol,Azure,TadGAN,Dense-AE,LSTM-DT,DONUT,LSTM-AE,TSI-GAN,},
    nodes near coords,
    xlabel= \% Improvement,
]
\addplot[fill=cyan] coordinates {(13.04,TSI-GAN)(-14.25,LSTM-AE)(-15.21,DONUT)(-22.7,LSTM-DT)(-61.59,Dense-AE)
(-68.35,TadGAN)(-87.92,Azure)(-88.16,Luminol)};
\draw[black, ultra thick] (axis cs: 0,)--(axis cs: 0,0);

\end{axis}
\end{tikzpicture}}
\caption{Comparing all anomaly detection methods against MERLIN in terms of F1 score averaged across all 250 datasets, expressed as a percentage of improvement.}
\label{fig:percent_imp}
\vspace{-20pt}
\end{wrapfigure}

\begin{figure*}[t]

\begin{subfigure}[]{\linewidth}
    \includegraphics[width=\textwidth]{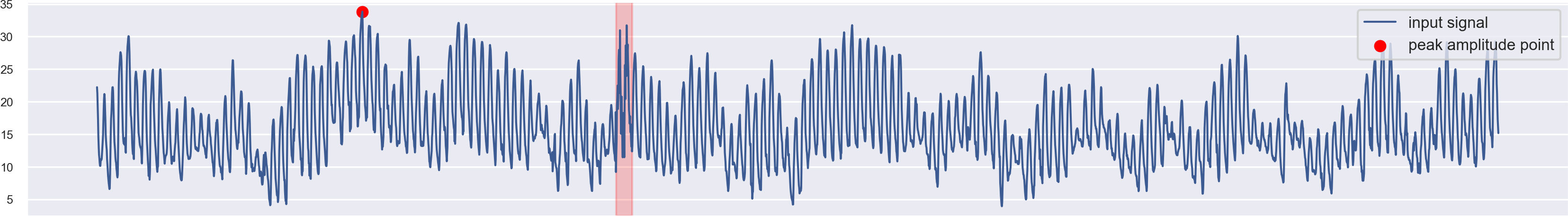}
    \label{fig:air_temp_other}
    \end{subfigure}
    
\begin{subfigure}[]{\linewidth}
\includegraphics[width=\textwidth]{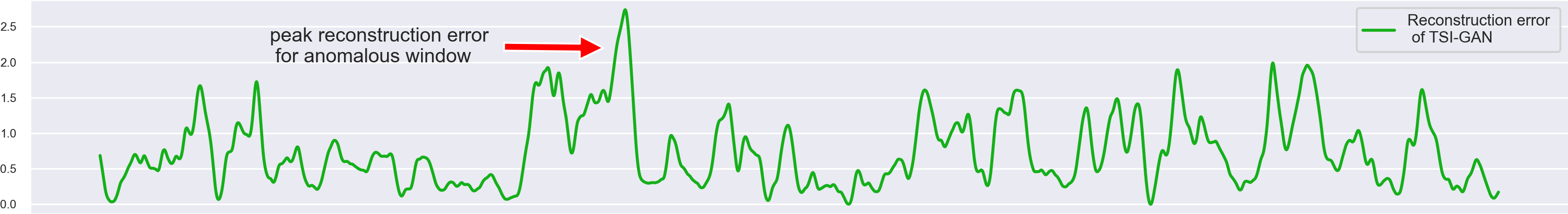}
\label{fig:proposed_rec}
\end{subfigure}

\begin{subfigure}[]{\linewidth}
\includegraphics[width=\textwidth]{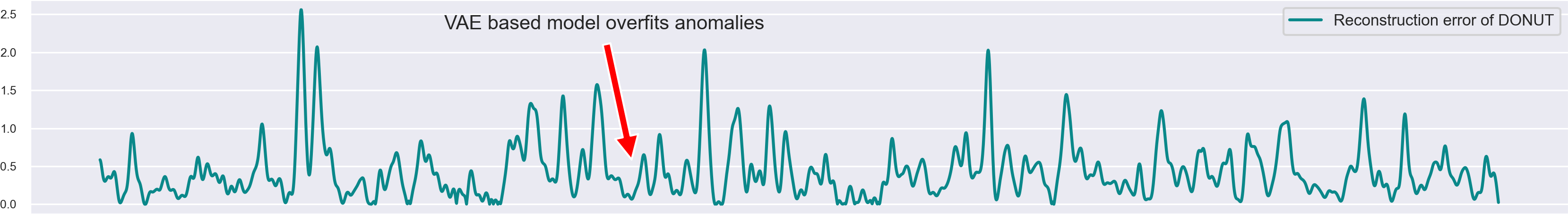}
\label{fig:merlin_rec}
\end{subfigure}

\begin{subfigure}[]{\linewidth}
\includegraphics[width=\textwidth]{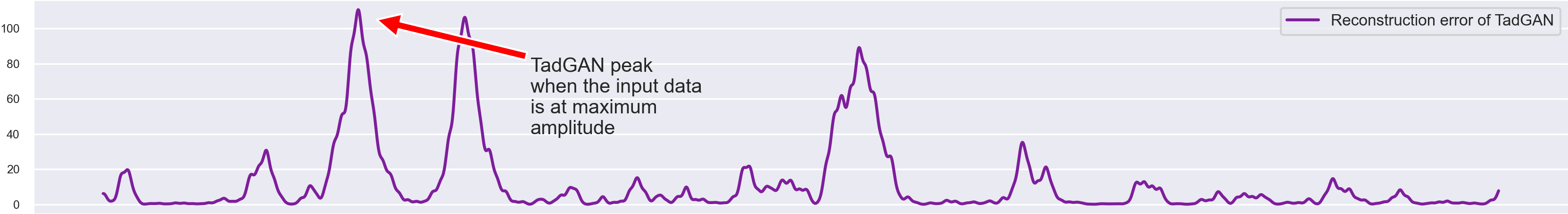}
\label{fig:tadgan_rec}
\end{subfigure}

\caption{Illustration of TSI-GAN vs. DONUT vs. TadGAN when applied to an example time series. The translucent red interval depicts the ground-truth anomaly.}
\label{fig:dl_based_methods}
\vspace{-24pt}
\end{figure*}

Overall, it is observed that TSI-GAN achieves an F1-Score of 0.468 and Precision of 0.445, outperforming all the baseline methods. More specifically, TSI-GAN offers an improvement of 13\% and 31\% on F1-score over the second and the third best methods, MERLIN (0.414) and LSTM-AE (0.355), respectively. We note that 85-95\% of the improvement was attributed to GAN, while 5-15\% was attributed to post-processing.  When the individual categories are considered, TSI-GAN performs the best on AirTemperature, PowerDemand, EPG for both original and distorted datasets and wins over other methods by a significant margin; it also offers competitive performance on other categories (InternalBleeding, NASA-T1 Distorted, and Noise) as well.

Using MERLIN as a benchmark, we measure the performance difference between each method and MERLIN in \fref{fig:percent_imp}. It indicates that TSI-GAN is the only one that offers a positive performance improvement while all the other methods underperform MERLIN.



Among all the deep learning-based methods, LSTM-AE performs the best, with an average F1-Score of 0.355; DONUT comes in second with a slightly lower score of 0.351. We examine the possible reasons for their shortfall as compared to TSI-GAN and how our approach overcomes them. As we mentioned earlier, autoencoder-based methods carry the risk of overfitting anomalies during training, by reconstructing anomalous samples just as accurately as normal samples. DONUT which employs VAE has this tendency as can be seen in \fref{fig:dl_based_methods}. This is also a plausible reason for the underperformance of other autoencoder-based models such as LSTM-AE.

In contrast, TSI-GAN uses an adversarial training strategy which makes our model largely immune to this behavior. However, while TadGAN and many others alike also use adversarial learning, they are unable to capture anomalies that are not amplitude spikes or dips unless dataset-specific parameters such as sampling interval are known. The reason is that they do not instrument feature engineering to capture anomalies that deviate in seasonality, trend, etc., and therefore tend to only detect extremely high or low amplitude points in the input as can be observed in \fref{fig:dl_based_methods}. It is the main reason why TadGAN only performs well on datasets in which all anomalies are either amplitude spikes or dips; such anomalies, however, are {\em trivial} to detect as pointed out by \cite{wu2021current}. On the other hand, TSI-GAN uses GAF and RP encoding which substantially enhances its ability to detect various types of non-trivial deviance, as can be observed in \fref{fig:dl_based_methods}.


\textbf{Time Efficiency.} We report the training and inference time of TSI-GAN in Table \ref{tab:efficiency of TSI-GAN}. The times are measured on a NVIDIA RTX 3070 GPU with 8GB of VRAM along with AMD Ryzen 7 5800H @ 3.20 GHz CPU. 


\begin{table*}[t]
\resizebox{0.95\textwidth}{!} {
\centering
\begin{tabular}{|c|ccc|ccc|}
\hline
                   & \multicolumn{3}{c|}{\textbf{AirTemperature}}                                           & \multicolumn{3}{c|}{\textbf{PowerDemand}}                                              \\ \hline
                   & \multicolumn{1}{c|}{\# of Samples} & \multicolumn{1}{c|}{Total time} & Per-window time & \multicolumn{1}{c|}{\# of Samples} & \multicolumn{1}{c|}{Total time} & Per-window time \\ \hline
\textbf{Training}  & \multicolumn{1}{c|}{7996}          & \multicolumn{1}{c|}{251s}       & 0.06s           & \multicolumn{1}{c|}{29772}         & \multicolumn{1}{c|}{246s}       & 0.01s           \\ \hline
\textbf{Inference} & \multicolumn{1}{c|}{4083}          & \multicolumn{1}{c|}{10s}        & 0.002s          & \multicolumn{1}{c|}{11862}         & \multicolumn{1}{c|}{27s}        & 0.002s          \\ \hline
\end{tabular}
}
\caption{Training and Inference time of TSI-GAN}
\label{tab:efficiency of TSI-GAN}
\vspace{-30pt}
\end{table*}

We can see that the training time remains almost constant irrespective of the number of training samples, the reason is that we train for iterations and not epochs. More importantly, the inference time per window is only two milliseconds, which signifies that TSI-GAN is well suited for use on rapidly arriving streaming data.


\vspace{-14pt}
\section{Conclusion}
\vspace{-10pt}
In this paper, we introduce TSI-GAN, a novel convolutional cycle-consistent GAN architecture that learns to reconstruct 2D-encoded time-series data and produces effective and reliable reconstruction errors for detecting time series anomalies. We also address the challenge of mitigating false alarms by post-processing the reconstruction error using a filtering technique and computing a reliable score by combining two channels. Our extensive experimental results demonstrate that TSI-GAN outperforms 8 state-of-the-art baseline methods over 250 non-trivial datasets that are well-curated. We also provide an in-depth analysis of the baselines' limitations and how our model addresses them. TSI-GAN is unsupervised and generalizes well without the need for parameter calibration, enabling it to be applicable to many applications that involve time series. 

\vspace{-12pt}
%
%
%
\footnotesize{\bibliographystyle{splncs04}
\bibliography{bib-file}}
\end{document}